\documentclass[10pt,twocolumn]{article}

\usepackage[top=1in,bottom=1in,left=0.75in,right=0.75in,columnsep=0.25in]{geometry}
\usepackage{times}
\usepackage{amsmath,amssymb,amsfonts}
\usepackage{graphicx}
\usepackage{booktabs}
\usepackage{multirow}
\usepackage{hyperref}
\usepackage{xcolor}
\usepackage{caption}
\usepackage{subcaption}
\usepackage{float}
\usepackage{placeins}
\usepackage{microtype}
\usepackage{natbib}
\usepackage{array}
\usepackage{makecell}
\usepackage{url}
\usepackage{enumitem}
\usepackage{parskip}
\usepackage{setspace}
\usepackage{tikz}
\usetikzlibrary{arrows.meta, positioning, shapes.geometric}
\hypersetup{
  colorlinks=true,
  linkcolor=blue!70!black,
  citecolor=blue!70!black,
  urlcolor=blue!70!black,
}

\title{
  \vspace{-1em}
  \textbf{Predict and Reconstruct: Joint Objectives for\\[2pt]
  Self-Supervised Language Representation Learning}
  \vspace{-0.5em}
}

\author{
  Aimen Boukhari\\[2pt]
  \'Ecole Nationale Sup\'erieure d'Informatique (ESI), Algiers , Algeria\\
  \href{mailto:mn_boukhari@esi.dz}{mn\_boukhari@esi.dz}
}

\date{}

\begin{document}
\maketitle
\vspace{-1em}

\begin{abstract}
\noindent
Masked language modelling (MLM) has been the dominant pre-training
objective for text encoders since BERT, yet it encourages representations
that are strongly anchored to surface-form token identity rather than
deeper semantic structure.
Inspired by the success of Joint Embedding Predictive Architectures
(JEPA)~\citep{lecun2022path} in vision and audio, we propose a hybrid
pre-training objective that combines a JEPA-style latent-space prediction
loss with an MLM reconstruction loss over a single shared encoder.
A learnable scalar $\lambda$ continuously balances the two objectives
during training.
We pre-train both a hybrid model and a pure-MLM baseline on English
Wikipedia using identical architectures and compute budgets (NVIDIA H100).
Extensive representation analysis across five GLUE benchmarks (SST-2,
MRPC, MNLI, CoLA, STS-B) using four pooling strategies reveals that
the hybrid encoder produces significantly more uniform embeddings
(uniformity $\leq -0.16$ vs.\ $-0.05$ for MLM), exhibits richer spectral
geometry under max pooling, encodes less surface-level lexical information,
and achieves a better semantic-to-lexical balance.
Despite similar linear-probe downstream accuracy, the geometric
differences are consistent and significant, suggesting that the JEPA
predictive objective reshapes the latent space in ways that standard
accuracy metrics alone cannot capture.
Code and checkpoints: \url{https://github.com/aymen-000/predict-reconstruct-language-models}
\end{abstract}

\section{Introduction}
\label{sec:intro}

Self-supervised learning (SSL) has transformed representation learning
across modalities.
In computer vision, contrastive methods such as
SimCLR~\citep{chen2020simclr} and MoCo~\citep{he2020moco}, followed by
non-contrastive approaches like BYOL~\citep{grill2020byol} and
VICReg~\citep{bardes2022vicreg}, demonstrated that powerful visual
features can be learned without labels.
In speech and audio, wav2vec~\citep{baevski2020wav2vec} and
data2vec~\citep{baevski2022data2vec} showed that masked prediction in
latent space generalises across modalities.
These successes share a common principle: rather than reconstructing
pixels or waveforms, the model learns to \emph{predict abstract
representations} of masked or future content.

Yann LeCun formalised this intuition in~\citep{lecun2022path},
arguing that reconstruction-based objectives are fundamentally limited
because they force the model to allocate capacity to irrelevant
low-level details.
The Joint Embedding Predictive Architecture (JEPA) avoids this by
having a predictor network match the \emph{target encoder's latent
representation} of the masked region, never reconstructing tokens
in pixel or token space.
The image instantiation I-JEPA~\citep{assran2023ijepa} demonstrated on
ImageNet that JEPA pre-training produces representations that generalise
better with fewer labelled examples than masked autoencoders, and that
learned features are more semantically structured as evidenced by linear
probing.

In natural language processing, BERT~\citep{devlin2019bert} established
MLM as the standard SSL objective.
While highly effective, MLM operates in token space: the model must
predict the exact identity of masked tokens, incentivising the encoder
to retain fine-grained lexical information at the expense of broader
semantic structure.
Several analyses have confirmed that BERT representations are highly
contextualised yet remain sensitive to surface-form
variation~\citep{ethayarajh2019contextual,rogers2020primer}.
SimCSE~\citep{gao2021simcse} and DeCLUTR~\citep{giorgi2021declutri}
improved the \emph{uniformity} of BERT-derived sentence embeddings
through contrastive fine-tuning, confirming that the MLM objective alone
does not fully exploit the embedding hypersphere.

A natural question therefore arises: can a JEPA-style latent prediction
objective be combined with MLM to produce text encoders that encode
semantics more robustly?
A recent independent line of work, LLM-JEPA~\citep{huangllmjepa},
explores applying JEPA principles to autoregressive language models.
Our work is complementary: we study the \emph{representation geometry}
of small-to-medium encoders trained under hybrid versus pure-MLM
objectives, providing the first systematic analysis using
alignment/uniformity metrics, eigenspectrum analysis, effective rank,
and probing classifiers across multiple GLUE tasks.

\paragraph{Contributions.}
\begin{enumerate}[leftmargin=*,topsep=2pt,itemsep=0pt,parsep=1pt]
  \item We propose a hybrid pre-training architecture that jointly
        optimises a JEPA cosine prediction loss and an MLM cross-entropy
        loss through a single shared encoder with a learnable balance
        weight~$\lambda$.
  \item We conduct the first systematic geometric analysis of hybrid
        vs.\ MLM-only text encoders across five GLUE datasets and four
        pooling strategies using six complementary representation metrics.
  \item We show that hybrid training consistently improves embedding
        uniformity and spectral richness while reducing surface-form
        bias, even under a small pre-training budget.
  \item We release code and model checkpoints to facilitate reproduction
        and extension of this analysis.
\end{enumerate}

\section{Background and Related Work}
\label{sec:background}

\subsection{Reconstruction-Based SSL and Its Limitations}

Masked language modelling, introduced in BERT~\citep{devlin2019bert},
predicts masked tokens from context.
This objective is effective but carries an implicit bias: the model must
memorise token-level statistics to recover the correct token identity,
encouraging representations to retain lexical surface form rather than
semantic content.
\citet{lecun2022path} identifies reconstruction objectives as
fundamentally misaligned with the goal of learning abstract world
models: predicting every detail of the input wastes model capacity on
unpredictable or irrelevant information.
In vision, masked autoencoders (MAE)~\citep{he2022mae} achieve strong
results but require fine-tuning to match JEPA-style methods on linear
evaluation~\citep{assran2023ijepa}, consistent with the hypothesis that
pixel reconstruction does not optimally produce semantic features.

\subsection{JEPA and Latent Prediction}

I-JEPA~\citep{assran2023ijepa} instantiates LeCun's JEPA framework for
images: a context encoder processes visible patches, a predictor maps
context representations to target representations, and the target
encoder (updated via exponential moving average, EMA) produces
representations of masked regions that the predictor must match.
The loss is computed in representation space rather than pixel space,
avoiding the reconstruction trap.
V-JEPA~\citep{bardes2024vjepa} extends this to video and
MC-JEPA~\citep{bardes2023mcjepa} adds motion consistency.
LLM-JEPA~\citep{huangllmjepa} recently proposed adapting JEPA to
autoregressive language models.
Our work differs in focus: we study encoder representation geometry
rather than downstream generation quality, and provide direct comparison
with a controlled MLM baseline.

\subsection{Representation Quality Metrics}

\citet{wang2020alignment} introduced \emph{alignment} and
\emph{uniformity} for evaluating sentence embeddings on the
$\ell_2$-normalised hypersphere.
\citet{roy2007effective} defined the \emph{effective rank} as the
exponential of the Shannon entropy of the normalised singular-value
distribution.
\citet{vershynin2018highDim} introduces the \emph{stable rank} as a
robust alternative.
\citet{garrido2023duality} used eigenspectrum analysis to compare
contrastive and non-contrastive SSL objectives.
\citet{conneau2018senteval} established the probing task framework for
analysing what linguistic information is encoded in sentence
representations, and \citet{ethayarajh2019contextual} applied
contextuality analysis to BERT representations.

\section{Method}
\label{sec:method}

\subsection{Architecture Overview}

\begin{figure}[t]
\centering
\includegraphics[width=0.95\linewidth]{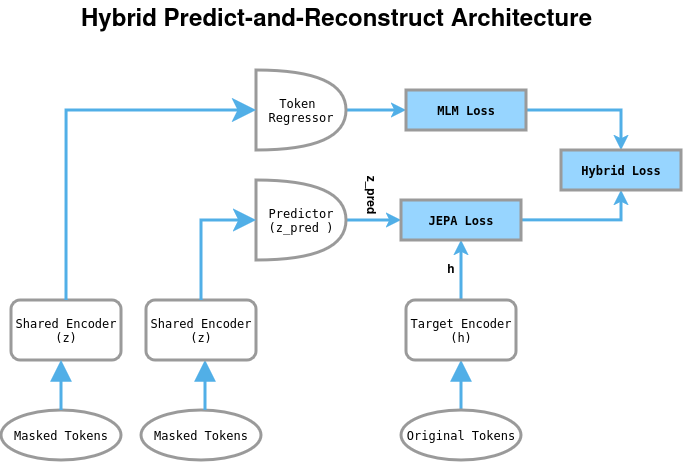}
\caption{Hybrid Predict-and-Reconstruct architecture. The shared encoder
produces contextual representations used by two branches: a predictor
for the JEPA objective and a token regression head for the MLM
objective. The target encoder is updated using exponential moving
average (EMA) and provides stable targets for representation
prediction.}
\label{fig:hybrid_arch}
\end{figure}

Our architecture consists of three components: a shared encoder
$f_\theta$, a predictor $g_\phi$, and a target encoder $\bar{f}_\theta$
updated via EMA.
The shared encoder processes input tokens and produces contextualised
representations used for both the JEPA and MLM objectives.
A lightweight token regressor $h_\psi$ maps encoder outputs to
vocabulary logits for the MLM branch.
The overall architecture is illustrated in Figure~\ref{fig:hybrid_arch}.

\subsection{Hybrid Pre-training Objective}

Given a token sequence $\mathbf{x}$ of length $L$, we apply two
distinct masking operations.

\paragraph{Block masking (JEPA branch).}
Following I-JEPA~\citep{assran2023ijepa}, we sample contiguous block
masks.
An encoder mask $\mathcal{M}_\text{enc}$ (scale $[0.65, 0.85]$) defines
the visible context; prediction masks
$\{\mathcal{M}_\text{pred}^k\}_{k=1}^{2}$ (scale $[0.10, 0.25]$)
define the target regions.

\paragraph{BERT masking (MLM branch).}
We apply standard BERT-style masking with probability $p=0.15$:
80\% of selected positions receive the \texttt{[MASK]} token,
10\% are replaced by a random token, and 10\% are left unchanged.

\paragraph{Forward pass.}
\begin{align}
\mathbf{z} &= f_\theta(\tilde{\mathbf{x}},\,\mathcal{M}_\text{enc})
  \label{eq:z}\\
\hat{\mathbf{h}} &= g_\phi(\mathbf{z},\,\mathcal{M}_\text{enc},\,
                   \mathcal{M}_\text{pred})
  \label{eq:zhat}\\
\mathbf{h} &= \bar{f}_\theta(\mathbf{x})
  \quad\text{(no grad)}
  \label{eq:h}\\
\mathbf{z}_\text{full} &= f_\theta(\tilde{\mathbf{x}})
  \label{eq:zfull}
\end{align}
where $\tilde{\mathbf{x}}$ denotes the BERT-masked token sequence,
$\mathbf{h}$ is the target representation, and $\mathbf{z}_\text{full}$
is the full-sequence latent used for token regression.

\paragraph{Loss functions.}
\begin{align}
\mathcal{L}_\text{JEPA}
  &= 1 - \frac{1}{|\mathcal{B}|}\sum_{(i,j)\in\mathcal{B}}
    \cos\!\bigl(\hat{\mathbf{h}}_{ij},\,\mathbf{h}_{ij}\bigr)
    \label{eq:ljepa}\\
\mathcal{L}_\text{MLM}
  &= \mathrm{CE}\!\bigl(h_\psi(\mathbf{z}_\text{full}),\,
    \mathbf{y}_\text{mask}\bigr)
    \label{eq:lmlm}\\
\mathcal{L}
  &= \lambda\,\mathcal{L}_\text{JEPA}
    + (1-\lambda)\,\mathcal{L}_\text{MLM},\quad
    \lambda = \sigma(w)
    \label{eq:total}
\end{align}
where $w \in \mathbb{R}$ is a learnable scalar optimised end-to-end and
$\sigma$ is the sigmoid function.

\paragraph{EMA update.}
\begin{equation}
  \bar{\boldsymbol{\theta}} \leftarrow
  m\,\bar{\boldsymbol{\theta}} + (1-m)\,\boldsymbol{\theta},
  \quad m \in [0.996,\,1.0]
  \label{eq:ema}
\end{equation}

\subsection{Token Regressor}

The token regressor $h_\psi$ operates directly on the shared encoder's
latent representation:
\begin{equation}
  h_\psi(\mathbf{z}) =
  W_2\,\mathrm{LN}\!\bigl(\mathrm{GELU}(W_1 \mathbf{z})\bigr)
  \label{eq:head}
\end{equation}
with $W_1 \in \mathbb{R}^{D\times D}$ ,
$W_2 \in \mathbb{R}^{D\times V}$ ($V$ = vocabulary size)
and $\mathrm{LN}(\cdot)$ denotes layer normalisation~\citep{ba2016layer}.Unlike a standard MLM head, $h_\psi$ receives gradients from both
objectives through the shared encoder weights.

\subsection{Sentence Pooling}

For downstream evaluation, token representations are aggregated via
mean pooling over non-padding positions:
\begin{equation}
  \mathbf{s} = \frac{\sum_{t=1}^{L}
    \mathbf{1}[x_t \neq \texttt{pad}]\,\mathbf{z}_t}
    {\sum_{t=1}^{L} \mathbf{1}[x_t \neq \texttt{pad}]}
  \label{eq:pool}
\end{equation}
This matches the pooling used in all linear-probe fine-tuning experiments.

\section{Experimental Setup}
\label{sec:experiments}

\subsection{Pre-training}

\paragraph{Data.}
Both models are pre-trained on English Wikipedia
(WikiText).
All text is tokenised with the \texttt{bert-base-uncased} tokeniser
(vocabulary size 30,522) and sequences are truncated or padded to
512~tokens.

\paragraph{Model architecture.}
The model uses a transformer-based encoder with token embedding dimension $d_{\text{emb}} = 512$. The predictor consists of 6 layers with embedding dimension $d_{\text{pred}} = 512$. A target encoder, with the same architecture as the main encoder, is maintained with frozen weights and updated via exponential moving average. The token regressor head maps the encoder outputs back to the vocabulary ($\text{vocab\_size} = 30{,}522$) for masked language modeling. Optimization is performed with AdamW, using a learnable scalar to balance the JEPA (cosine) and MLM (cross-entropy) losses.

\paragraph{Training.}
Both models are trained for 3~epochs with batch size~64 on a single
NVIDIA H100 GPU using bfloat16 mixed precision and AdamW with cosine
learning rate schedule (peak LR $5\times10^{-5}$, weight decay~$0.05$).
Full hyperparameters are given in Table~\ref{tab:config}.

\begin{table}[h]
\centering\small
\caption{Pre-training hyperparameters.}
\label{tab:config}
\setlength{\tabcolsep}{5pt}
\begin{tabular}{lc}
\toprule
\textbf{Hyperparameter} & \textbf{Value} \\
\midrule
Epochs              & 3 \\
Batch size          & 64 \\
Peak LR             & $5\times10^{-5}$ \\
LR schedule         & cosine \\
Warmup steps        & 10 \\
Weight decay        & 0.05 \\
Precision           & bfloat16 \\
EMA range           & [0.996,\;1.0] \\
Encoder mask scale  & [0.65,\;0.85] \\
Pred.\ mask scale   & [0.10,\;0.25] \\
Num.\ pred.\ masks  & 2 \\
Min.\ keep tokens   & 32 \\
MLM probability     & 0.15 \\
Hardware            & NVIDIA H100 \\
\bottomrule
\end{tabular}
\end{table}

\subsection{Downstream Evaluation: Linear Probing}

We evaluate frozen encoder representations via linear probing on five
GLUE tasks~\citep{wang2018glue}.
The encoder is kept frozen throughout; only a lightweight head is trained.
Since GLUE test labels are not publicly available, all results are
reported on the official validation splits.
No hyperparameter search was performed on the validation set; the same
configuration was applied to both models.

\paragraph{Single-sentence tasks (SST-2, CoLA).}
Mean-pooled representations feed a
LayerNorm\;$\to$\;Dropout(0.2)\;$\to$\;Linear head.

\paragraph{Sentence-pair tasks (MRPC, MNLI).}
Both sentences are encoded independently; the classifier receives
$[\mathbf{s}_1;\;\mathbf{s}_2;\;|\mathbf{s}_1-\mathbf{s}_2|;\;
\mathbf{s}_1\odot\mathbf{s}_2]$~\citep{conneau2017supervised}.

\paragraph{Regression task (STS-B).}
Cosine similarity is calibrated to the $[0,5]$ score range via a
learnable scale and bias:
$\hat{y} = \sigma_w \cdot \cos(\mathbf{s}_1,\mathbf{s}_2) + b_w$.

Per-task fine-tuning details are given in Table~\ref{tab:finetuning}.

\begin{table}[h]
\centering\small
\caption{Fine-tuning hyperparameters per task.}
\label{tab:finetuning}
\setlength{\tabcolsep}{5pt}
\begin{tabular}{lccc}
\toprule
\textbf{Task} & \textbf{LR} & \textbf{Epochs} & \textbf{Metric} \\
\midrule
SST-2 & $1\times10^{-3}$ & 90 & Accuracy \\
MRPC  & $1\times10^{-3}$ & 10 & F1 \\
MNLI  & $1\times10^{-3}$ & 15 & Acc.\ (matched) \\
STS-B & $1\times10^{-3}$ & 15 & Spearman $\rho$ \\
\bottomrule
\end{tabular}
\end{table}

\subsection{Representation Analysis}

We extract frozen sentence embeddings for up to 2,000 samples from each
task's validation split and compute six metrics:

\begin{enumerate}[leftmargin=*,topsep=2pt,itemsep=0pt,parsep=1pt]
  \item \textbf{Spectral entropy}~\citep{garrido2023duality}:
        $H_\text{spec} = H(\boldsymbol{\sigma}/\|\boldsymbol{\sigma}\|_1)
        / \log D$.
  \item \textbf{Effective rank}~\citep{roy2007effective}:
        $\text{erank} = \exp H(\boldsymbol{\sigma}/\|\boldsymbol{\sigma}\|_1)$.
  \item \textbf{Stable rank}~\citep{vershynin2018highDim}:
        $\text{srank} = \|\mathbf{Z}\|_F^2 / \|\mathbf{Z}\|_2^2$.
  \item \textbf{Alignment}~\citep{wang2020alignment}:
        mean squared $\ell_2$ distance between same-class pairs.
  \item \textbf{Uniformity}~\citep{wang2020alignment}:
        $\log \mathbb{E}[e^{-2\|\mathbf{u}-\mathbf{v}\|^2}]$ on
        $\ell_2$-normalised embeddings.
  \item \textbf{Probe gap}~\citep{conneau2018senteval}:
        semantic probe accuracy minus token probe accuracy.
\end{enumerate}

All metrics are computed under four pooling strategies: mean, max,
weighted mean, and attention pooling.
Full formal definitions are given in Appendix~\ref{app:metric_definitions}.

\section{Results}
\label{sec:results}

\subsection{Downstream Task Accuracy}

Table~\ref{tab:downstream} reports linear-probe performance on five
GLUE tasks.
Both models achieve comparable accuracy, consistent with the known
finding that MLM baselines are strong linear classifiers under
mean-pooled representations~\citep{gao2021simcse}.

\begin{table}[h]
\centering\small
\caption{Linear-probe downstream performance on GLUE validation splits.
Frozen encoder; best per-task result in \textbf{bold}.
Spearman $\rho$ for STS-B.}
\label{tab:downstream}
\setlength{\tabcolsep}{6pt}
\begin{tabular}{lcc}
\toprule
\textbf{Task} & \textbf{Hybrid} & \textbf{MLM-only} \\
\midrule
SST-2\;(Acc.)         & 67.55        & \textbf{68.69} \\
MRPC\;(F1)            & \textbf{63.09} & 59.84 \\
MNLI\;(Acc.)          & 50.82        & \textbf{51.36} \\
STS-B$^\dagger$\;(Spearman) & 0.281 & \textbf{0.283} \\
\bottomrule
\end{tabular}
\end{table}

\subsection{Representation Geometry}

\subsubsection{Uniformity}

The hybrid encoder consistently achieves significantly more negative
uniformity scores across all five datasets and all four pooling
strategies (Table~\ref{tab:uniformity}).
The mean uniformity under attention pooling is $-0.54$ for hybrid
vs.\ $-0.07$ for MLM-only — a sevenfold difference.
This confirms that the JEPA predictive objective prevents
representational collapse and promotes a more isotropic use of the
embedding hypersphere~\citep{wang2020alignment}, analogous to findings
in vision~\citep{assran2023ijepa}.

\begin{table}[h]
\centering\small
\caption{Uniformity ($\downarrow$ better) by dataset and pooling.}
\label{tab:uniformity}
\setlength{\tabcolsep}{4pt}
\begin{tabular}{llcc}
\toprule
\textbf{Dataset} & \textbf{Pooling} & \textbf{Hybrid} & \textbf{MLM} \\
\midrule
\multirow{4}{*}{SST-2}
  & mean      & $\mathbf{-0.160}$ & $-0.052$ \\
  & max       & $\mathbf{-0.294}$ & $-0.090$ \\
  & weighted  & $\mathbf{-0.160}$ & $-0.052$ \\
  & attention & $\mathbf{-0.448}$ & $-0.055$ \\
\midrule
\multirow{4}{*}{MRPC}
  & mean      & $\mathbf{-0.134}$ & $-0.053$ \\
  & max       & $\mathbf{-0.262}$ & $-0.088$ \\
  & weighted  & $\mathbf{-0.134}$ & $-0.053$ \\
  & attention & $\mathbf{-0.269}$ & $-0.055$ \\
\midrule
\multirow{4}{*}{MNLI}
  & mean      & $\mathbf{-0.169}$ & $-0.063$ \\
  & max       & $\mathbf{-0.290}$ & $-0.096$ \\
  & weighted  & $\mathbf{-0.163}$ & $-0.063$ \\
  & attention & $\mathbf{-0.365}$ & $-0.067$ \\
\midrule
\multirow{4}{*}{CoLA}
  & mean      & $\mathbf{-0.314}$ & $-0.079$ \\
  & max       & $\mathbf{-0.365}$ & $-0.098$ \\
  & weighted  & $\mathbf{-0.314}$ & $-0.079$ \\
  & attention & $\mathbf{-0.955}$ & $-0.083$ \\
\midrule
\multirow{4}{*}{STS-B}
  & mean      & $\mathbf{-0.202}$ & $-0.067$ \\
  & max       & $\mathbf{-0.321}$ & $-0.102$ \\
  & weighted  & $\mathbf{-0.201}$ & $-0.068$ \\
  & attention & $\mathbf{-0.577}$ & $-0.069$ \\
\bottomrule
\end{tabular}
\end{table}

\subsubsection{Alignment--Uniformity Trade-off}

The improved uniformity comes at the cost of higher within-class
alignment values for hybrid representations.
MLM-only alignment is consistently near zero ($\leq 0.002$), indicating
extremely tight class clusters, while hybrid alignment is larger
(e.g.\ SST-2/attention: 0.20 vs.\ 0.001), reflecting a more relaxed
intra-class structure.
This alignment-uniformity trade-off~\citep{wang2020alignment} directly
explains why linear-probe accuracy is similar for both models: linear
classifiers benefit primarily from tight clusters, which favours
MLM-only representations.
Scatter plots of alignment vs.\ uniformity and intra/inter-class
distance ratios for all datasets are given in
Appendix~\ref{app:alignment_uniformity}.

\subsubsection{Spectral Analysis}

Under max pooling, the hybrid encoder achieves higher spectral entropy
and effective rank across all datasets (Table~\ref{tab:spectral}),
indicating that more embedding dimensions carry meaningful variance.
The stable rank is also consistently higher for hybrid under max
pooling.

\begin{table}[h]
\centering\small
\caption{Spectral metrics under max pooling. H\,=\,Hybrid; M\,=\,MLM-only.
Eff.\ Rank~\citep{roy2007effective}; Srank~\citep{vershynin2018highDim}.}
\label{tab:spectral}
\setlength{\tabcolsep}{3.5pt}
\begin{tabular}{lcccccc}
\toprule
 & \multicolumn{2}{c}{\textbf{Spec.\ Ent.}\;($\uparrow$)}
 & \multicolumn{2}{c}{\textbf{Eff.\ Rank}\;($\uparrow$)}
 & \multicolumn{2}{c}{\textbf{Srank}\;($\uparrow$)} \\
\cmidrule(lr){2-3}\cmidrule(lr){4-5}\cmidrule(lr){6-7}
\textbf{Task} & H & M & H & M & H & M \\
\midrule
SST-2 & \textbf{0.765} & 0.735 & \textbf{384} & 364 & \textbf{6.98} & 4.11 \\
MRPC  & 0.794 & \textbf{0.824} & 299 & \textbf{301} & 8.18 & \textbf{11.33} \\
MNLI  & \textbf{0.779} & 0.767 & \textbf{406} & 391 & \textbf{8.35} & 4.99 \\
CoLA  & \textbf{0.752} & 0.715 & \textbf{365} & 330 & \textbf{10.82} & 4.75 \\
STS-B & \textbf{0.743} & 0.721 & \textbf{390} & 372 & \textbf{6.94} & 4.09 \\
\bottomrule
\end{tabular}
\end{table}

\subsubsection{Probing Results}

\paragraph{Token probe.}
The hybrid encoder consistently exhibits lower token probe accuracy
(Table~\ref{tab:probing}), indicating that its representations retain
less surface-level lexical information.
Across all tasks under attention pooling, hybrid token accuracy is
3--9 points lower than MLM-only.

\paragraph{Semantic probe.}
Despite encoding less token-level information, hybrid representations
achieve comparable or slightly higher semantic probe performance.
Under max pooling on MRPC, hybrid achieves the only positive probe gap
($+0.031$), confirming that the JEPA objective promotes semantic
encoding over lexical memorisation.

\paragraph{CoLA MCC.}
The hybrid encoder achieves consistently positive MCC on CoLA
($0.052$--$0.064$ vs.\ $-0.023$--$0.010$ for MLM), suggesting that
block-span prediction induces a weak but reproducible syntactic
sensitivity absent from token-level MLM.

\begin{table}[h]
\centering\small
\caption{Probing results under attention pooling.
H\,=\,Hybrid; M\,=\,MLM-only.}
\label{tab:probing}
\setlength{\tabcolsep}{3pt}
\begin{tabular}{lcccccc}
\toprule
 & \multicolumn{2}{c}{\textbf{Token}\;($\downarrow$)}
 & \multicolumn{2}{c}{\textbf{Semantic}\;($\uparrow$)}
 & \multicolumn{2}{c}{\textbf{Gap}\;($\uparrow$)} \\
\cmidrule(lr){2-3}\cmidrule(lr){4-5}\cmidrule(lr){6-7}
\textbf{Task} & H & M & H & M & H & M \\
\midrule
SST-2 & \textbf{0.663} & 0.754 & \textbf{0.643} & 0.635 & \textbf{$-$0.019} & $-$0.119 \\
MRPC  & \textbf{0.616} & 0.678 & 0.608 & \textbf{0.610} & \textbf{$-$0.008} & $-$0.068 \\
MNLI  & \textbf{0.685} & 0.740 & 0.326 & \textbf{0.333} & \textbf{$-$0.359} & $-$0.407 \\
CoLA  & \textbf{0.813} & 0.870 & \textbf{0.602} & 0.594 & \textbf{$-$0.211} & $-$0.276 \\
STS-B & \textbf{0.592} & 0.654 & 0.187 & \textbf{0.193} & \textbf{$-$0.405} & $-$0.460 \\
\bottomrule
\end{tabular}
\end{table}

\section{Ablation Study: Pooling Strategies}
\label{sec:ablation}

We evaluate four pooling strategies to disentangle the effect of
aggregation from the effect of the pre-training objective.

\paragraph{Mean pooling} (Eq.~\ref{eq:pool}) is the default and matches
the pooling used during fine-tuning.
It provides the fairest comparison for downstream accuracy but yields
the most compressed spectral geometry.

\paragraph{Max pooling} consistently yields the highest spectral entropy
and effective rank for both models.
It is the most discriminative strategy for the hybrid encoder on
classification tasks, and the only configuration under which the probe
gap turns positive for MRPC.

\paragraph{Weighted mean pooling} produces results nearly identical to
mean pooling across all metrics.
Since weights are initialised uniformly and evaluated inside
\texttt{no\_grad}, this serves as a useful null result confirming that
observed geometric differences are intrinsic to the encoder rather than
to the pooling arithmetic.

\paragraph{Attention pooling} amplifies differences between the two
models more than any other strategy.
The uniformity gap is widest under attention pooling
(SST-2: $-0.448$ vs.\ $-0.055$; STS-B: $-0.577$ vs.\ $-0.069$),
suggesting that the JEPA objective most strongly affects
high-attention positions --- the tokens the model considers most
semantically salient.

\begin{figure}[h]
  \centering
  \includegraphics[width=\linewidth]{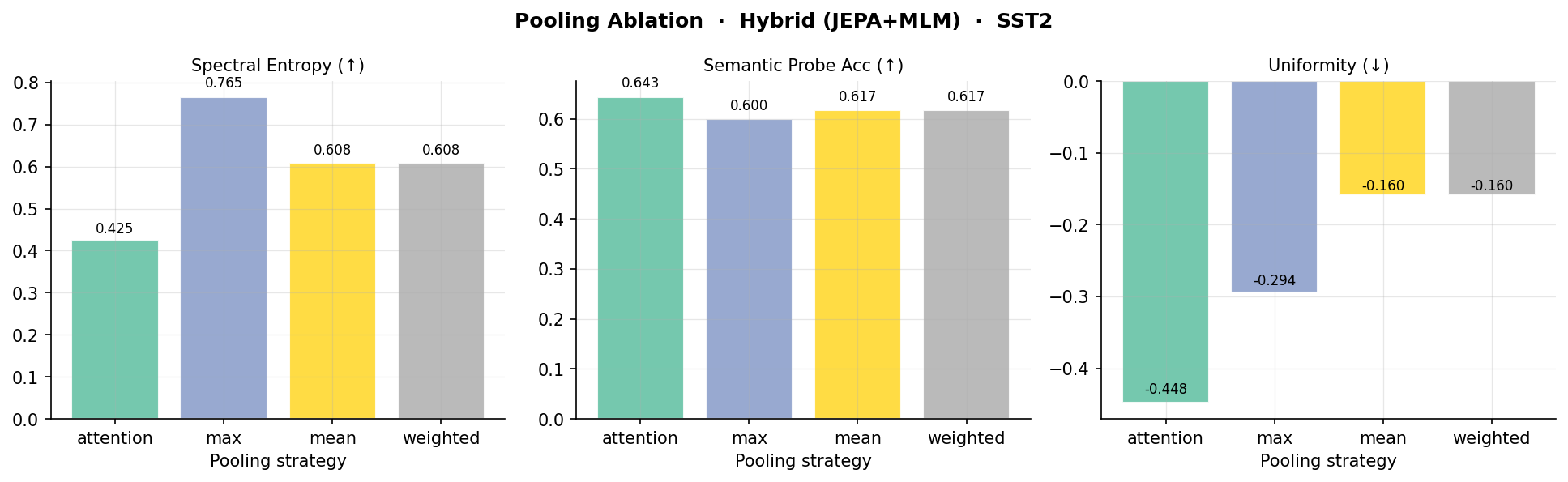}
  \caption{Spectral entropy, semantic probe accuracy, and uniformity for
  the hybrid encoder under four pooling strategies on SST-2.
  Max pooling maximises spectral richness; attention pooling maximises
  the uniformity advantage.}
  \label{fig:pooling_ablation}
\end{figure}

Figure~\ref{fig:pooling_ablation} illustrates the pooling ablation for
the hybrid encoder on SST-2.

\section{Discussion}
\label{sec:discussion}

\paragraph{Why similar accuracy despite different geometry?}
The alignment-uniformity trade-off~\citep{wang2020alignment} provides
the explanation.
Hybrid representations cover the embedding hypersphere more uniformly
but do not cluster same-class points as tightly.
Since linear classifiers benefit primarily from tight clusters, MLM's
alignment advantage offsets hybrid's uniformity advantage at the
linear-probe evaluation level.
We expect the uniformity advantage to translate to downstream gains
under non-linear probing, retrieval tasks, or longer pre-training.

\paragraph{Attention pooling as a diagnostic tool.}
The amplified differences under attention pooling suggest that JEPA's
effect is concentrated at salient positions rather than distributed
uniformly across the sequence, consistent with the block-masking design
that forces the predictor to recover representations of contiguous spans.

\paragraph{Resource constraints.}
Pre-training was intentionally limited to 3~epochs on a modest corpus
to study the effect of the objective under a controlled compute budget.
The geometric differences observed are therefore likely a lower bound
on what is achievable at scale.

\section{Future Work}
\label{sec:future}

\begin{itemize}[leftmargin=*,topsep=2pt,itemsep=0pt,parsep=2pt]
  \item \textbf{Alternative prediction losses.}
        Cosine similarity enforces directional alignment but ignores
        magnitude.
        Future work should evaluate smooth $\ell_1$,
        VICReg~\citep{bardes2022vicreg}, or Barlow
        Twins~\citep{zbontar2021barlow} losses in the JEPA branch.
  \item \textbf{Curriculum-based $\lambda$.}
        The current $\lambda$ converges slowly.
        A schedule starting MLM-heavy and shifting to JEPA-heavy as the
        target encoder matures may accelerate learning.
  \item \textbf{Alternative masking strategies.}
        Span masking~\citep{joshi2020spanbert}, whole-word masking, and
        syntactically informed masking are natural alternatives.
  \item \textbf{Scale.}
        Extending to BookCorpus\,+\,Wikipedia, more epochs, and larger
        model sizes is the most direct path to assessing whether
        geometric advantages translate to accuracy improvements.
  \item \textbf{Non-linear probing and retrieval.}
        Evaluating with MLP probes and semantic similarity benchmarks
        (STS12--STS16, SICK-R) would provide a more complete picture.
  \item \textbf{Collapse monitoring.}
        Online tracking of uniformity and effective rank during training
        could serve as an adaptive objective-switching signal.
\end{itemize}

\section{Conclusion}
\label{sec:conclusion}

We proposed a hybrid pre-training objective that combines a JEPA-style
latent prediction loss with masked language modelling over a single
shared encoder.
Through systematic representation analysis across five GLUE benchmarks
and four pooling strategies, we showed that the hybrid objective
consistently produces more uniform embedding distributions, richer
spectral geometry, and better semantic-to-lexical balance compared to a
controlled MLM-only baseline trained under identical conditions.
These advantages are not captured by linear-probe accuracy alone,
highlighting the value of geometric representation analysis as a
complementary evaluation protocol.
Our findings provide empirical support for LeCun's hypothesis that
latent-space prediction objectives lead to more abstract representations
than token reconstruction~\citep{lecun2022path}, and constitute a step toward understanding how JEPA principles can be applied to language encoders under realistic resource constraints.

\bibliographystyle{plainnat}
\bibliography{references}

@inproceedings{assran2023ijepa,
  author = {Assran, M. and Duval, Q. and Misra, I. and Bojanowski, P. and Vincent, P. and Rabbat, M. and LeCun, Y. and Ballas, N.},
  title = {Self-supervised learning from images with a joint-embedding predictive architecture},
  booktitle = {CVPR 2023},
  year = {2023}
}

@inproceedings{baevski2020wav2vec,
  author = {Baevski, A. and Zhou, H. and Mohamed, A. and Auli, M.},
  title = {wav2vec 2.0: A framework for self-supervised learning of speech representations},
  booktitle = {NeurIPS 2020},
  year = {2020}
}

@inproceedings{baevski2022data2vec,
  author = {Baevski, A. and Hsu, W.-N. and Xu, Q. and Babu, A. and Gu, J. and Auli, M.},
  title = {data2vec: A general framework for self-supervised learning in speech, vision and language},
  booktitle = {ICML 2022},
  year = {2022}
}

@inproceedings{bardes2022vicreg,
  author = {Bardes, A. and Ponce, J. and LeCun, Y.},
  title = {VICReg: Variance-invariance-covariance regularization for self-supervised learning},
  booktitle = {ICLR 2022},
  year = {2022}
}

@article{bardes2023mcjepa,
  author = {Bardes, A. and Ponce, J. and LeCun, Y.},
  title = {MC-JEPA: A Joint-Embedding Predictive Architecture for Self-Supervised Learning of Motion and Content Features},
  journal = {arXiv preprint arXiv:2307.12698},
  year = {2023}
}

@article{ba2016layer,
  author = {Ba, J.L. and Kiros, J.R. and Hinton, G.E.},
  title = {Layer normalization},
  journal = {arXiv:1607.06450},
  year = {2016}
}

@article{bardes2024vjepa,
  author = {Bardes, A. and Garrido, Q. and Ponce, J. and Rabbat, M. and LeCun, Y. and Assran, M. and Ballas, N.},
  title = {Revisiting feature prediction for learning visual representations from video},
  journal = {arXiv:2404.08471},
  year = {2024}
}

@inproceedings{chen2020simclr,
  author = {Chen, T. and Kornblith, S. and Norouzi, M. and Hinton, G.},
  title = {A simple framework for contrastive learning of visual representations},
  booktitle = {ICML 2020},
  year = {2020}
}

@inproceedings{conneau2018senteval,
  author = {Conneau, A. and Kiela, D.},
  title = {SentEval: An evaluation toolkit for universal sentence representations},
  booktitle = {LREC 2018},
  year = {2018}
}

@inproceedings{conneau2017supervised,
  author = {Conneau, A. and Kiela, D. and Schwentz, H. and Barrault, L. and Bordes, A.},
  title = {Supervised learning of universal sentence representations from natural language inference data},
  booktitle = {EMNLP 2017},
  year = {2017}
}

@inproceedings{devlin2019bert,
  author = {Devlin, J. and Chang, M.-W. and Lee, K. and Toutanova, K.},
  title = {BERT: Pre-training of deep bidirectional transformers for language understanding},
  booktitle = {NAACL 2019},
  year = {2019}
}

@inproceedings{ethayarajh2019contextual,
  author = {Ethayarajh, K.},
  title = {How contextual are contextualized word representations? Comparing the geometry of BERT, ELMo, and GPT-2 embeddings},
  booktitle = {EMNLP 2019},
  year = {2019}
}

@inproceedings{gao2021simcse,
  author = {Gao, T. and Yao, X. and Chen, D.},
  title = {SimCSE: Simple contrastive learning of sentence embeddings},
  booktitle = {EMNLP 2021},
  year = {2021}
}

@inproceedings{garrido2023duality,
  author = {Garrido, Q. and Chen, Y. and Bardes, A. and Najman, L. and LeCun, Y.},
  title = {On the duality between contrastive and non-contrastive self-supervised learning},
  booktitle = {ICLR 2023},
  year = {2023}
}

@inproceedings{giorgi2021declutri,
  author = {Giorgi, J. and Nitski, O. and Wang, B. and Bader, G.},
  title = {DeCLUTR: Deep contrastive learning for unsupervised textual representations},
  booktitle = {ACL 2021},
  year = {2021}
}

@inproceedings{grill2020byol,
  author = {Grill, J.-B. and Strub, F. and Altch\'e, F. and others},
  title = {Bootstrap your own latent: A new approach to self-supervised learning},
  booktitle = {NeurIPS 2020},
  year = {2020}
}

@article{huangllmjepa,
  author = {Huang, H. and LeCun, Y. and Balestriero, R.},
  title = {LLM-JEPA: Large Language Models Meet Joint Embedding Predictive Architectures},
  journal = {arXiv preprint arXiv:2509.14252},
  year = {2025}
}

@inproceedings{he2020moco,
  author = {He, K. and Fan, H. and Wu, Y. and Xie, S. and Girshick, R.},
  title = {Momentum contrast for unsupervised visual representation learning},
  booktitle = {CVPR 2020},
  year = {2020}
}

@inproceedings{he2022mae,
  author = {He, K. and Chen, X. and Xie, S. and Li, Y. and Doll\'ar, P. and Girshick, R.},
  title = {Masked autoencoders are scalable vision learners},
  booktitle = {CVPR 2022},
  year = {2022}
}

@inproceedings{joshi2020spanbert,
  author = {Joshi, M. and Chen, D. and Liu, Y. and Weld, D.S. and Zettlemoyer, L. and Levy, O.},
  title = {SpanBERT: Improving pre-training by representing and predicting spans},
  journal = {TACL 2020},
  year = {2020}
}

@misc{lecun2022path,
  author = {LeCun, Y.},
  title = {A path towards autonomous machine intelligence},
  howpublished = {OpenReview preprint},
  year = {2022}
}

@article{rogers2020primer,
  author = {Rogers, A. and Kovaleva, O. and Rumshisky, A.},
  title = {A primer in BERTology: What we know about how BERT works},
  journal = {TACL 2020},
  year = {2020}
}

@inproceedings{roy2007effective,
  author = {Roy, O. and Vetterli, M.},
  title = {The effective rank: A measure of effective dimensionality},
  booktitle = {EUSIPCO 2007},
  year = {2007}
}

@book{vershynin2018highDim,
  author = {Vershynin, R.},
  title = {High-Dimensional Probability},
  publisher = {Cambridge University Press},
  year = {2018}
}

@inproceedings{wang2020alignment,
  author = {Wang, T. and Isola, P.},
  title = {Understanding contrastive representation learning through alignment and uniformity on the hypersphere},
  booktitle = {ICML 2020},
  year = {2020}
}

@inproceedings{wang2018glue,
  author = {Wang, A. and Singh, A. and Michael, J. and Hill, F. and Levy, O. and Bowman, S.},
  title = {GLUE: A multi-task benchmark and analysis platform for natural language understanding},
  booktitle = {ICLR 2019},
  year = {2018}
}

@inproceedings{zbontar2021barlow,
  author = {Zbontar, J. and Jing, L. and Misra, I. and LeCun, Y. and Deny, S.},
  title = {Barlow twins: Self-supervised learning via redundancy reduction},
  booktitle = {ICML 2021},
  year = {2021}
}

\clearpage
\onecolumn

\appendix
\section{Full Representation Analysis Tables}
\label{app:full_tables}

Tables~\ref{tab:full_sst2}--\ref{tab:full_stsb} report all six
representation metrics for all pooling strategies across all datasets.
H\,=\,Hybrid; M\,=\,MLM-only.

\begin{table}[H]
\centering\small
\caption{Full representation metrics --- SST-2.}
\label{tab:full_sst2}
\setlength{\tabcolsep}{4pt}
\begin{tabular}{lcccccccccccc}
\toprule
& \multicolumn{2}{c}{\textbf{Spec.\ Ent.}}
  & \multicolumn{2}{c}{\textbf{Eff.\ Rank}}
  & \multicolumn{2}{c}{\textbf{Srank}}
  & \multicolumn{2}{c}{\textbf{Uniformity}}
  & \multicolumn{2}{c}{\textbf{Sem.\ Probe}}
  & \multicolumn{2}{c}{\textbf{Probe Gap}} \\
\cmidrule(lr){2-3}\cmidrule(lr){4-5}\cmidrule(lr){6-7}
\cmidrule(lr){8-9}\cmidrule(lr){10-11}\cmidrule(lr){12-13}
\textbf{Pool} & H & M & H & M & H & M & H & M & H & M & H & M \\
\midrule
mean      & \textbf{0.609} & 0.605 & 242 & 249 & 4.49 & 4.50          & $\mathbf{-0.160}$ & $-$0.052 & 0.617 & 0.618          & $\mathbf{-0.138}$ & $-$0.152 \\
max       & \textbf{0.765} & 0.735 & \textbf{384} & 364 & \textbf{6.98} & 4.11 & $\mathbf{-0.294}$ & $-$0.090 & \textbf{0.600} & 0.595 & $\mathbf{-0.008}$ & $-$0.045 \\
weighted  & \textbf{0.609} & 0.605 & 242 & 249 & 4.49 & 4.50          & $\mathbf{-0.160}$ & $-$0.052 & 0.617 & 0.619          & $\mathbf{-0.139}$ & $-$0.149 \\
attention & 0.425 & 0.602          & 231 & 255 & 1.81 & 4.15          & $\mathbf{-0.448}$ & $-$0.055 & \textbf{0.643} & 0.635 & $\mathbf{-0.019}$ & $-$0.119 \\
\bottomrule
\end{tabular}
\end{table}

\begin{table}[H]
\centering\small
\caption{Full representation metrics --- MRPC.}
\label{tab:full_mrpc}
\setlength{\tabcolsep}{4pt}
\begin{tabular}{lcccccccccccc}
\toprule
& \multicolumn{2}{c}{\textbf{Spec.\ Ent.}}
  & \multicolumn{2}{c}{\textbf{Eff.\ Rank}}
  & \multicolumn{2}{c}{\textbf{Srank}}
  & \multicolumn{2}{c}{\textbf{Uniformity}}
  & \multicolumn{2}{c}{\textbf{Sem.\ Probe}}
  & \multicolumn{2}{c}{\textbf{Probe Gap}} \\
\cmidrule(lr){2-3}\cmidrule(lr){4-5}\cmidrule(lr){6-7}
\cmidrule(lr){8-9}\cmidrule(lr){10-11}\cmidrule(lr){12-13}
\textbf{Pool} & H & M & H & M & H & M & H & M & H & M & H & M \\
\midrule
mean      & \textbf{0.631} & 0.629 & 184 & 186 & 7.39 & 7.94          & $\mathbf{-0.134}$ & $-$0.053 & 0.583 & 0.635          & $-$0.052 & $-$0.038 \\
max       & 0.794 & \textbf{0.824} & 299 & 301 & 8.18 & 11.33         & $\mathbf{-0.262}$ & $-$0.088 & \textbf{0.618} & 0.576 & $\mathbf{+0.031}$ & $-$0.002 \\
weighted  & \textbf{0.631} & 0.629 & 184 & 186 & 7.39 & 7.94          & $\mathbf{-0.134}$ & $-$0.053 & 0.583 & 0.635          & $-$0.057 & $-$0.038 \\
attention & 0.611 & \textbf{0.642} & \textbf{198} & 194 & 4.13 & 8.73 & $\mathbf{-0.269}$ & $-$0.055 & 0.608 & \textbf{0.610} & $\mathbf{-0.008}$ & $-$0.068 \\
\bottomrule
\end{tabular}
\end{table}

\begin{table}[H]
\centering\small
\caption{Full representation metrics --- MNLI.}
\label{tab:full_mnli}
\setlength{\tabcolsep}{4pt}
\begin{tabular}{lcccccccccccc}
\toprule
& \multicolumn{2}{c}{\textbf{Spec.\ Ent.}}
  & \multicolumn{2}{c}{\textbf{Eff.\ Rank}}
  & \multicolumn{2}{c}{\textbf{Srank}}
  & \multicolumn{2}{c}{\textbf{Uniformity}}
  & \multicolumn{2}{c}{\textbf{Sem.\ Probe}}
  & \multicolumn{2}{c}{\textbf{Probe Gap}} \\
\cmidrule(lr){2-3}\cmidrule(lr){4-5}\cmidrule(lr){6-7}
\cmidrule(lr){8-9}\cmidrule(lr){10-11}\cmidrule(lr){12-13}
\textbf{Pool} & H & M & H & M & H & M & H & M & H & M & H & M \\
\midrule
mean      & \textbf{0.607} & 0.597 & 242 & 245          & \textbf{6.06} & 5.83 & $\mathbf{-0.169}$ & $-$0.063 & \textbf{0.362} & 0.354 & $\mathbf{-0.392}$ & $-$0.410 \\
max       & \textbf{0.779} & 0.767 & \textbf{406} & 391 & \textbf{8.35} & 4.99 & $\mathbf{-0.290}$ & $-$0.096 & 0.315 & \textbf{0.349} & $-$0.310 & $-$0.288 \\
weighted  & \textbf{0.607} & 0.597 & 242 & 245          & \textbf{6.06} & 5.83 & $\mathbf{-0.163}$ & $-$0.063 & \textbf{0.363} & 0.355 & $\mathbf{-0.392}$ & $-$0.409 \\
attention & 0.538 & \textbf{0.603} & 253 & 253          & 2.76 & \textbf{5.91} & $\mathbf{-0.365}$ & $-$0.067 & 0.326 & \textbf{0.333} & $-$0.359 & $-$0.407 \\
\bottomrule
\end{tabular}
\end{table}

\begin{table}[H]
\centering\small
\caption{Full representation metrics --- CoLA. Semantic probe reports MCC.}
\label{tab:full_cola}
\setlength{\tabcolsep}{4pt}
\begin{tabular}{lcccccccccccc}
\toprule
& \multicolumn{2}{c}{\textbf{Spec.\ Ent.}}
  & \multicolumn{2}{c}{\textbf{Eff.\ Rank}}
  & \multicolumn{2}{c}{\textbf{Srank}}
  & \multicolumn{2}{c}{\textbf{Uniformity}}
  & \multicolumn{2}{c}{\textbf{MCC}\;($\uparrow$)}
  & \multicolumn{2}{c}{\textbf{Probe Gap}} \\
\cmidrule(lr){2-3}\cmidrule(lr){4-5}\cmidrule(lr){6-7}
\cmidrule(lr){8-9}\cmidrule(lr){10-11}\cmidrule(lr){12-13}
\textbf{Pool} & H & M & H & M & H & M & H & M & H & M & H & M \\
\midrule
mean      & \textbf{0.606} & 0.586 & 235 & 239          & \textbf{5.16} & 4.58 & $\mathbf{-0.314}$ & $-$0.079 & \textbf{0.064} & 0.000   & $\mathbf{-0.246}$ & $-$0.284 \\
max       & \textbf{0.752} & 0.715 & \textbf{365} & 330 & \textbf{10.82} & 4.75 & $\mathbf{-0.365}$ & $-$0.098 & \textbf{0.024} & $-$0.023 & $\mathbf{-0.224}$ & $-$0.306 \\
weighted  & \textbf{0.606} & 0.586 & 235 & 239          & \textbf{5.16} & 4.58 & $\mathbf{-0.314}$ & $-$0.079 & \textbf{0.064} & 0.000   & $\mathbf{-0.246}$ & $-$0.285 \\
attention & 0.479 & \textbf{0.572} & 221 & \textbf{241} & 2.20 & \textbf{3.91} & $\mathbf{-0.955}$ & $-$0.083 & \textbf{0.052} & 0.010   & $\mathbf{-0.211}$ & $-$0.276 \\
\bottomrule
\end{tabular}
\end{table}

\begin{table}[H]
\centering\small
\caption{Full representation metrics --- STS-B. Semantic probe reports Spearman~$\rho$.}
\label{tab:full_stsb}
\setlength{\tabcolsep}{4pt}
\begin{tabular}{lcccccccccccc}
\toprule
& \multicolumn{2}{c}{\textbf{Spec.\ Ent.}}
  & \multicolumn{2}{c}{\textbf{Eff.\ Rank}}
  & \multicolumn{2}{c}{\textbf{Srank}}
  & \multicolumn{2}{c}{\textbf{Uniformity}}
  & \multicolumn{2}{c}{\textbf{Spearman}\;($\uparrow$)}
  & \multicolumn{2}{c}{\textbf{Probe Gap}} \\
\cmidrule(lr){2-3}\cmidrule(lr){4-5}\cmidrule(lr){6-7}
\cmidrule(lr){8-9}\cmidrule(lr){10-11}\cmidrule(lr){12-13}
\textbf{Pool} & H & M & H & M & H & M & H & M & H & M & H & M \\
\midrule
mean      & 0.581 & \textbf{0.588} & 228 & 235 & 4.83 & 5.53          & $\mathbf{-0.202}$ & $-$0.067 & \textbf{0.284} & 0.224 & $\mathbf{-0.369}$ & $-$0.449 \\
max       & \textbf{0.743} & 0.721 & \textbf{390} & 372 & \textbf{6.94} & 4.09 & $\mathbf{-0.321}$ & $-$0.102 & 0.140 & \textbf{0.167} & $\mathbf{-0.376}$ & $-$0.397 \\
weighted  & 0.581 & \textbf{0.588} & 227 & 235 & 4.83 & 5.53          & $\mathbf{-0.201}$ & $-$0.068 & \textbf{0.284} & 0.224 & $\mathbf{-0.369}$ & $-$0.446 \\
attention & 0.484 & \textbf{0.597} & 233 & \textbf{243} & 2.35 & \textbf{5.70} & $\mathbf{-0.577}$ & $-$0.069 & 0.187 & \textbf{0.193} & $\mathbf{-0.405}$ & $-$0.460 \\
\bottomrule
\end{tabular}
\end{table}

\FloatBarrier
\section{Alignment--Uniformity Plots}
\label{app:alignment_uniformity}

\begin{figure}[H]
  \centering
  \includegraphics[width=0.9\linewidth]{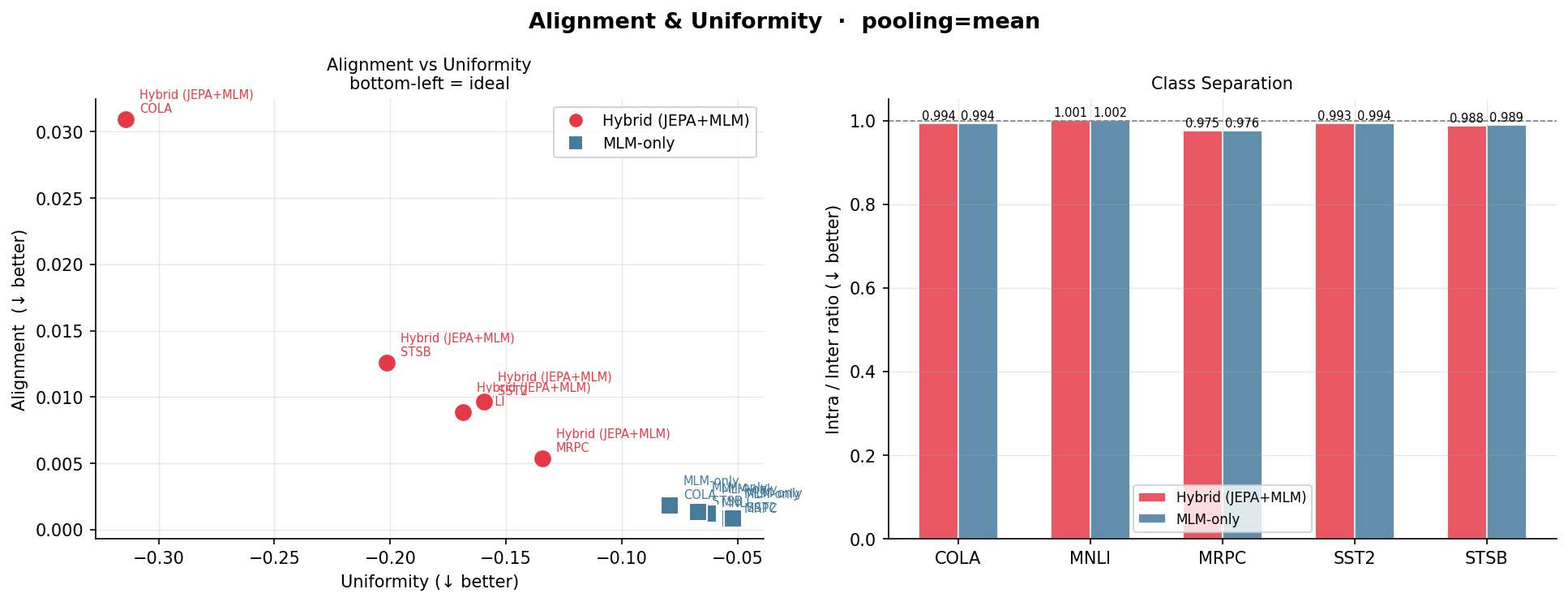}
  \caption{Alignment vs.\ uniformity and class separation --- mean pooling,
  all datasets.}
  \label{fig:au_sst2}
\end{figure}

\begin{figure}[H]
  \centering
  \includegraphics[width=0.9\linewidth]{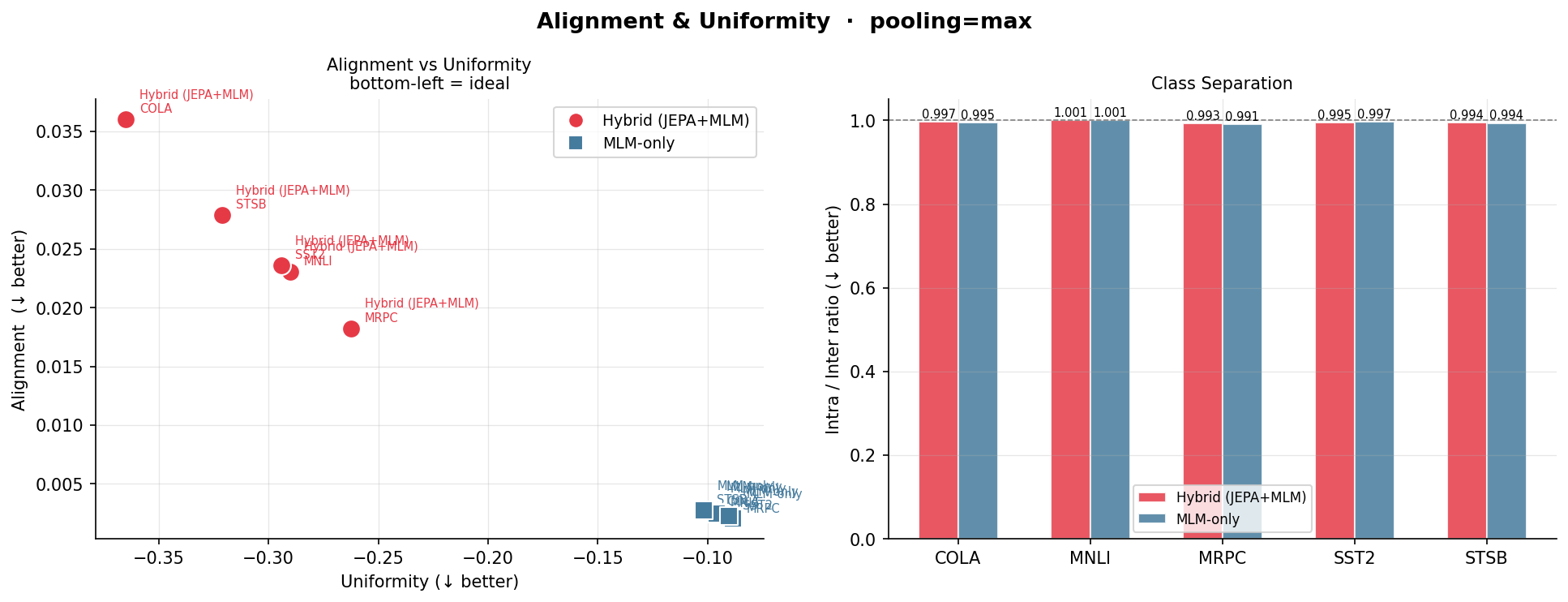}
  \caption{Alignment vs.\ uniformity and class separation --- max pooling,
  all datasets.}
  \label{fig:au_mrpc}
\end{figure}

\begin{figure}[H]
  \centering
  \includegraphics[width=0.9\linewidth]{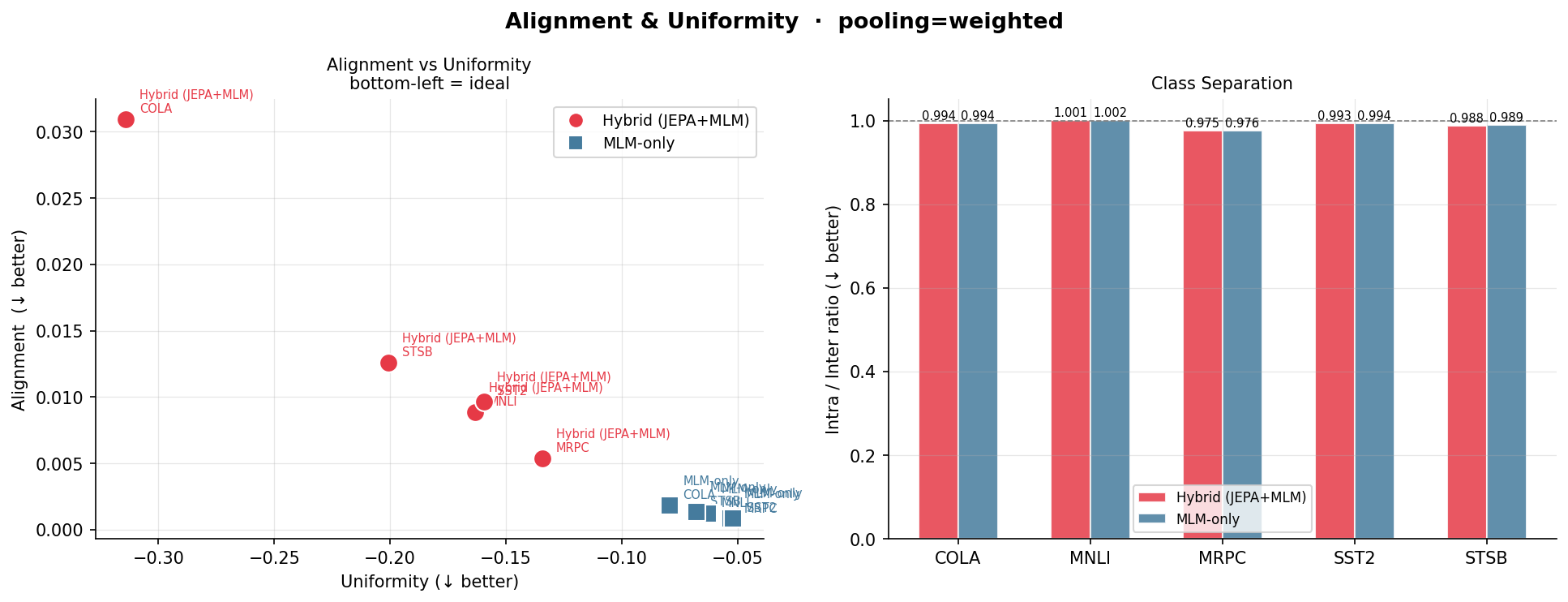}
  \caption{Alignment vs.\ uniformity and class separation --- weighted mean
  pooling, all datasets.}
  \label{fig:au_mnli}
\end{figure}

\begin{figure}[H]
  \centering
  \includegraphics[width=0.9\linewidth]{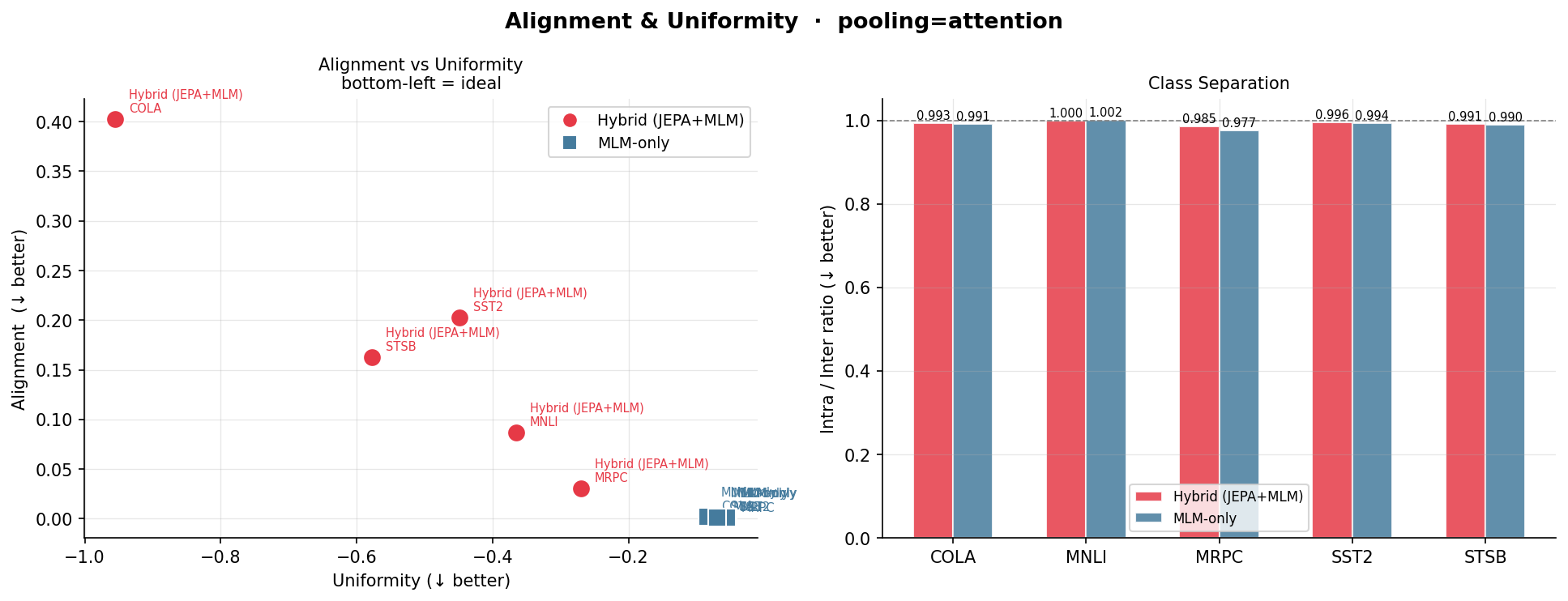}
  \caption{Alignment vs.\ uniformity and class separation --- attention
  pooling, all datasets.
  The uniformity gap between hybrid and MLM-only is most pronounced here,
  consistent with the quantitative results in Table~\ref{tab:uniformity}.}
  \label{fig:au_stsb}
\end{figure}

\FloatBarrier
\section{Eigenspectrum Plots}
\label{app:eigenspectrum}

\begin{figure}[H]
  \centering
  \includegraphics[width=0.9\linewidth]{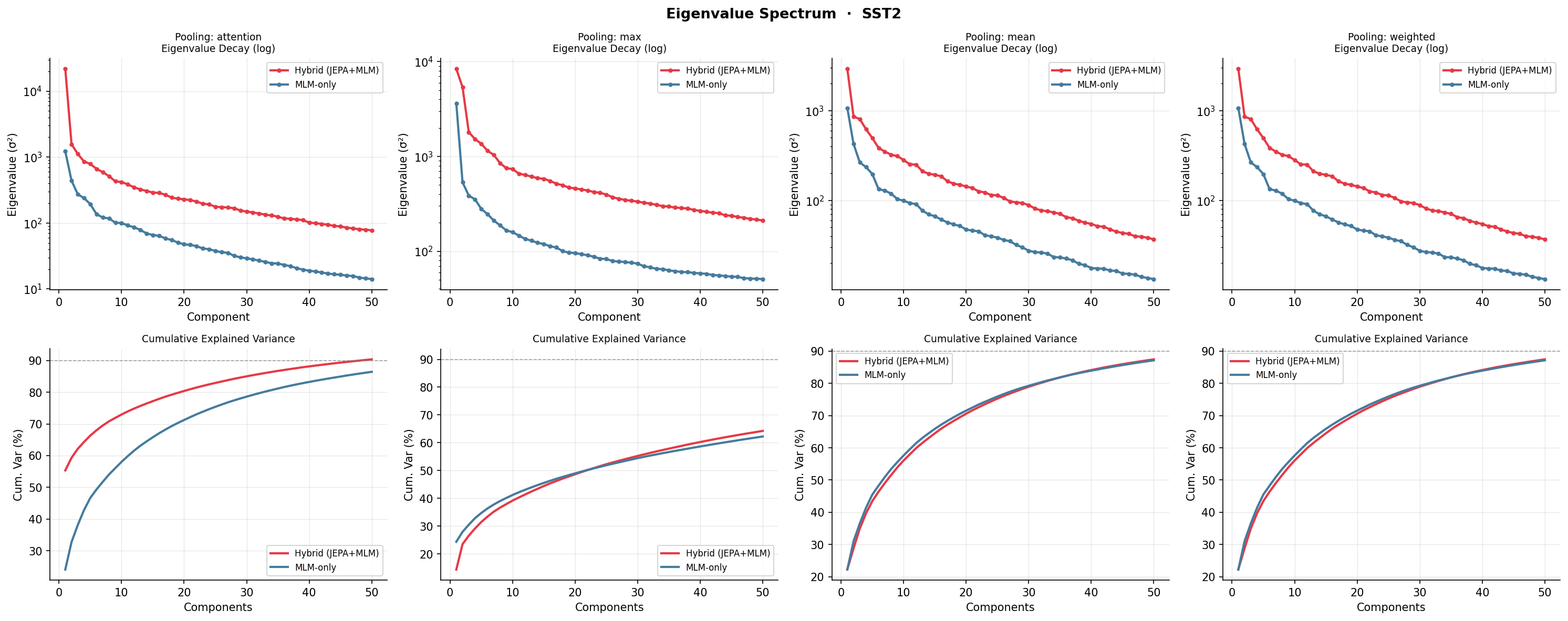}
  \caption{Eigenvalue spectrum --- SST-2, mean pooling.}
  \label{fig:eigen_sst2}
\end{figure}

\begin{figure}[H]
  \centering
  \includegraphics[width=0.9\linewidth]{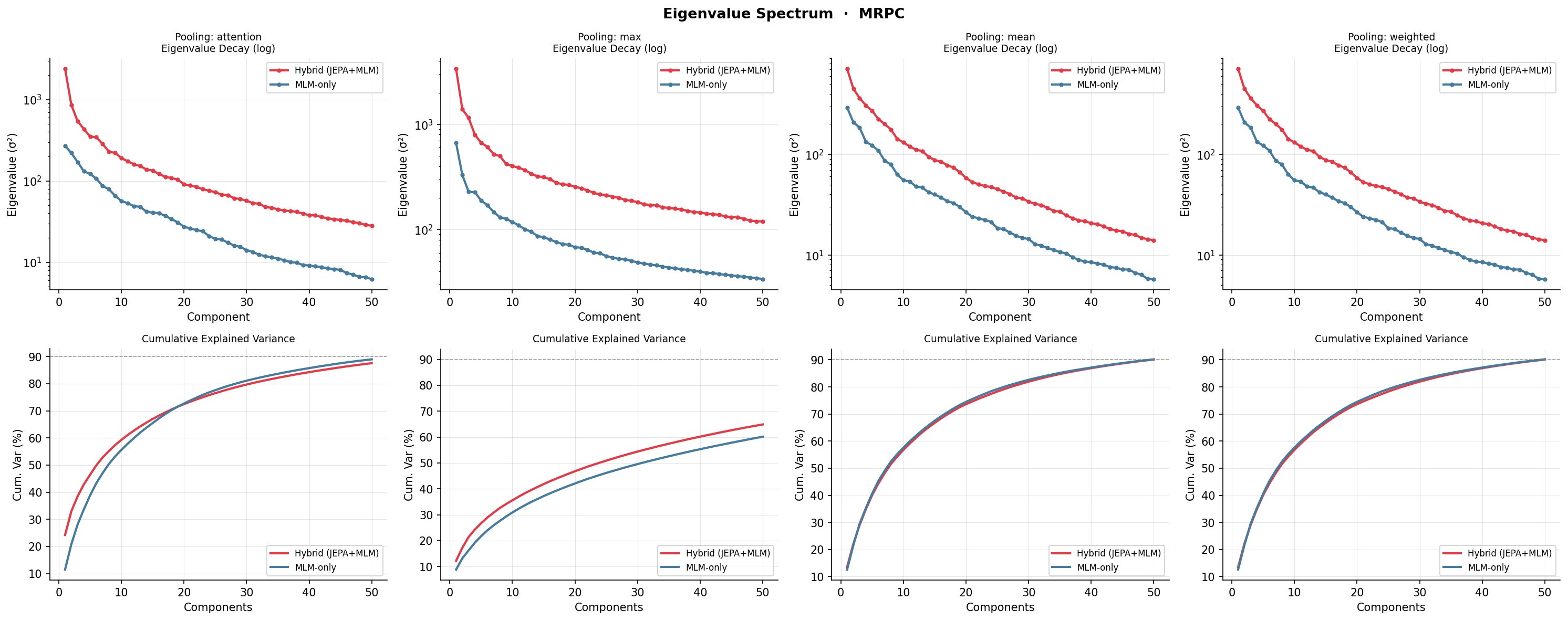}
  \caption{Eigenvalue spectrum --- MRPC, mean pooling.}
  \label{fig:eigen_mrpc}
\end{figure}

\begin{figure}[H]
  \centering
  \includegraphics[width=0.9\linewidth]{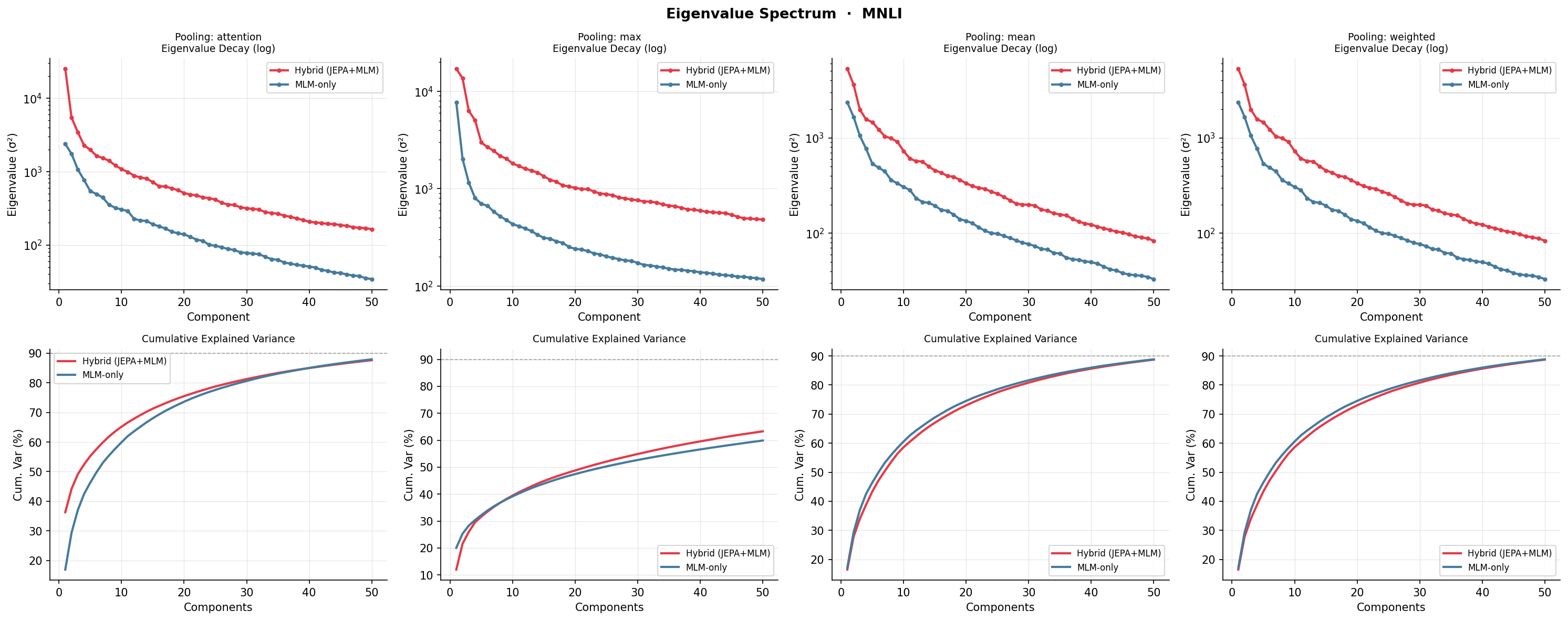}
  \caption{Eigenvalue spectrum --- MNLI, mean pooling.}
  \label{fig:eigen_mnli}
\end{figure}

\begin{figure}[H]
  \centering
  \includegraphics[width=0.9\linewidth]{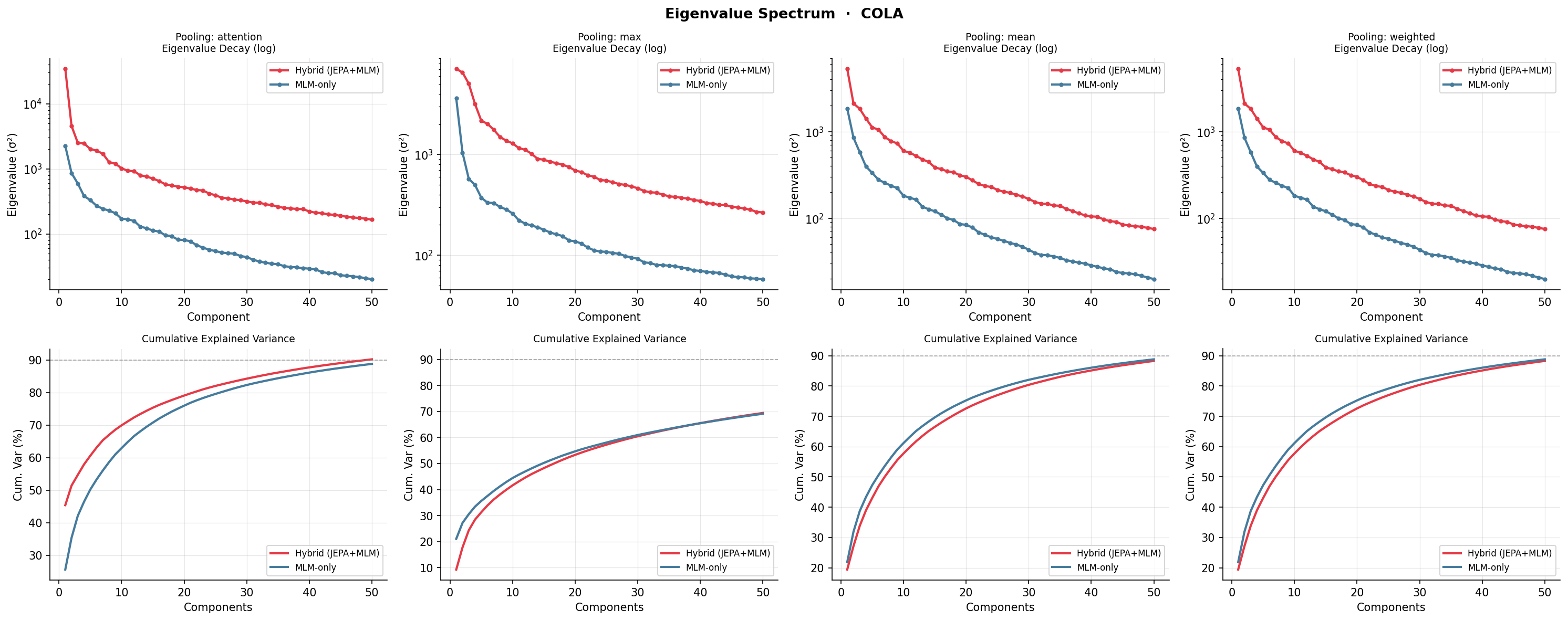}
  \caption{Eigenvalue spectrum --- CoLA, mean pooling.}
  \label{fig:eigen_cola}
\end{figure}

\FloatBarrier
\section{Representation Metric Definitions}
\label{app:metric_definitions}

Let $\mathbf{Z} \in \mathbb{R}^{N \times D}$ denote the centred embedding
matrix, and let $\boldsymbol{\sigma} = (\sigma_1, \ldots, \sigma_r)$ be
its singular values in descending order.
All metrics are computed on frozen sentence embeddings extracted from
the validation split.

\paragraph{1.\; Spectral Entropy~\citep{garrido2023duality}.}
\begin{equation}
  H_\text{spec} = \frac{H\!\left(\boldsymbol{\sigma}/
    \|\boldsymbol{\sigma}\|_1\right)}{\log D},
    \qquad
  H(p) = -\textstyle\sum_i p_i \log p_i
\end{equation}
$H_\text{spec} \in [0,1]$; a value of~1 means all singular values are
equal, indicating that every embedding dimension carries the same
variance (maximally rich representation).

\paragraph{2.\; Effective Rank~\citep{roy2007effective}.}
\begin{equation}
  \text{erank}(\mathbf{Z}) =
  \exp H\!\left(\boldsymbol{\sigma}/\|\boldsymbol{\sigma}\|_1\right)
\end{equation}
Measures the number of dimensions that effectively contribute to the
variance; higher values indicate richer representations.

\paragraph{3.\; Stable Rank~\citep{vershynin2018highDim}.}
\begin{equation}
  \text{srank}(\mathbf{Z}) =
  \frac{\|\mathbf{Z}\|_F^2}{\|\mathbf{Z}\|_2^2}
  = \frac{\sum_i \sigma_i^2}{\sigma_1^2}
\end{equation}
A robust alternative to effective rank that is less sensitive to
outlier singular values.

\paragraph{4.\; Alignment~\citep{wang2020alignment}.}
\begin{equation}
  \mathcal{A}(f;\,\alpha) =
  \underset{(\mathbf{u},\mathbf{v})\sim p_\text{pos}}{\mathbb{E}}
  \bigl[\|\hat{f}(\mathbf{u}) - \hat{f}(\mathbf{v})\|^2\bigr]^\alpha,
  \quad \alpha=2
\end{equation}
where $\hat{f}(\mathbf{x}) = f(\mathbf{x})/\|f(\mathbf{x})\|_2$ and
$p_\text{pos}$ is the distribution of same-class pairs.
Lower alignment means same-class representations cluster more tightly.

\paragraph{5.\; Uniformity~\citep{wang2020alignment}.}
\begin{equation}
  \mathcal{U}(f;\,t) =
  \log\,\underset{\mathbf{u},\mathbf{v}\overset{\text{i.i.d.}}{\sim}
    p_\text{data}}{\mathbb{E}}
  \bigl[e^{-t\|\hat{f}(\mathbf{u})-\hat{f}(\mathbf{v})\|^2}\bigr],
  \quad t=2
\end{equation}
More negative values indicate that embeddings are spread more uniformly
over the unit hypersphere, which prevents representational collapse.

\paragraph{6.\; Probe Gap~\citep{conneau2018senteval}.}
\begin{equation}
  \Delta_\text{probe} =
  \text{acc}_\text{semantic} - \text{acc}_\text{token}
\end{equation}
The semantic probe trains a linear classifier to predict the downstream
task label from the frozen embedding; the token probe predicts the
most frequent non-special token in the input sentence.
A positive gap indicates that the representation encodes task-relevant
semantics more strongly than surface-form lexical identity.

\end{document}